\documentclass[conference]{IEEEtran}
\IEEEoverridecommandlockouts

\usepackage{cite}
\usepackage{amsmath,amssymb,amsfonts,bm}
\usepackage[hyphens]{url}
\usepackage{graphicx}
\usepackage{textcomp}
\usepackage[dvipsnames,svgnames,x11names]{xcolor}
\usepackage{booktabs}
\usepackage{array,tabularx}
\usepackage{multirow}
\usepackage{colortbl}
\usepackage{tikz}
\usetikzlibrary{positioning,arrows.meta,calc,fit,backgrounds,shapes.geometric}
\usepackage[capitalize,nameinlink]{cleveref}
\usepackage{balance}

\definecolor{axonblue}{HTML}{1D4ED8}
\definecolor{axonblueLight}{HTML}{DBEAFE}
\definecolor{rawred}{HTML}{DC2626}
\definecolor{shapegreen}{HTML}{16A34A}
\definecolor{subpurple}{HTML}{7C3AED}
\definecolor{statorange}{HTML}{EA580C}
\definecolor{inkgray}{HTML}{374151}
\newcolumntype{Y}{>{\centering\arraybackslash}X}
\newcolumntype{Z}{>{\raggedright\arraybackslash}X}

\usepackage{pgfplots}
\pgfplotsset{compat=1.18}
\usetikzlibrary{arrows.meta, shapes.geometric}

\newcommand{\ExCal}{\textsc{ExCal}}
\newcommand{\R}{\mathbb{R}}
\newcommand{\clip}{\operatorname{clip}}
\newcommand{\ranknorm}{\operatorname{rank01}}
\newcommand{\uniformtable}{%
  \footnotesize
  \renewcommand{\arraystretch}{1.05}%
  \setlength{\tabcolsep}{2.5pt}%
}
\newcommand{\widetable}{%
  \scriptsize
  \renewcommand{\arraystretch}{1.04}%
  \setlength{\tabcolsep}{1.2pt}%
}
\AtBeginDocument{}

\makeatletter
\def\ps@IEEEtitlepagestyle{%
  \def\@oddfoot{\mycopyrightnotice}%
  \def\@evenfoot{}%
}

\def\mycopyrightnotice{%
  {\footnotesize
  \begin{minipage}{\textwidth}
  \centering
  \textcopyright~2026 IEEE. Personal use of this material is permitted.
  Permission from IEEE must be obtained for all other uses, in any current
  or future media, including reprinting/republishing this material for
  advertising or promotional purposes, creating new collective works,
  for resale or redistribution to servers or lists, or reuse of any
  copyrighted component of this work in other works.
  \end{minipage}}%
}
\makeatother

\begin{document}

\title{ExCal: Optimization-Free Multivariate Time Series Anomaly Detection via Excess Calibration}

\author{%
  Kadir\mbox{-}Kaan \"Ozer\IEEEauthorrefmark{1,}\IEEEauthorrefmark{2},
  Ren\'{e} Ebeling\IEEEauthorrefmark{1},
  Markus Enzweiler\IEEEauthorrefmark{2}%
  \thanks{%
    \IEEEauthorrefmark{1}Mercedes-Benz AG, Germany,
    \{kadir.oezer, rene.ebeling\}@mercedes-benz.com\par
    \IEEEauthorrefmark{2}Institute for Intelligent Systems,
    Esslingen University of Applied Sciences, Germany,
    \{markus.enzweiler\}@hs-esslingen.de%
  }%
}

\maketitle

\begin{abstract}
We introduce \ExCal{}, a normal-only multivariate time series anomaly detector that combines raw state retrieval, a signal/derivative subspace, and robust statistics across temporal scales. Each expert map receives a high normality floor from the training prefix. During scoring, positive excess is retained, attenuated for broadly firing maps, and combined with a small rank term. The method fits reference memories and projections without gradient optimization or anomaly labels. Prevalence and ranks depend on the complete scored sequence, making inference batch and transductive rather than online. On 180 series from 17 TSB-AD-M families, \ExCal{} obtains AUC-PR $0.421\pm0.37$ and VUS-PR $0.452\pm0.37$ over three seeds, ranking second among 24 methods. Holm-adjusted paired tests detect no significant difference from AxonAD or StreamVAE, which are the two leading non-pretrained methods on TSB-AD-M, on either primary metric, while finding significant positive differences against the other 21 baselines. A separate single-run probe over three series measures median model-wrapper time of $0.0662$ s. Ablations associate the largest primary-metric changes with raw retrieval and excess-magnitude preservation. The evaluated score is decomposable by expert, but its diagnostic utility and online suitability remain untested.
\end{abstract}


\section{Introduction}\label{sec:intro}

Multivariate telemetry may deviate through unusual operating points, broken cross-channel relations, changes in local dynamics, or events visible only at particular durations. With only a normal training prefix, no single representation is assured to expose all of these departures. Neural forecasting, reconstruction, and attention methods offer expressive alternatives~\cite{malhotra_lstm_2015,xu_unsupervised_2018,su_robust_2019,li_anomaly_2021,audibert_usad_2020,tuli_tranad_2022,xu_anomaly_2022,wu_timesnet_2023}, but require optimization and architecture choices. Classical approaches offer transparent normality tests based on isolation, density, projection, or retrieval~\cite{liu_isolation_2008,hariri_extended_2021,breunig_lof_2000,goldstein_hbos_2012,li_copod_2020,candes_robust_2011,cover_knn_1967,MACKIEWICZ1993303}. Combining these heterogeneous signals remains a calibration problem because their score scales differ.

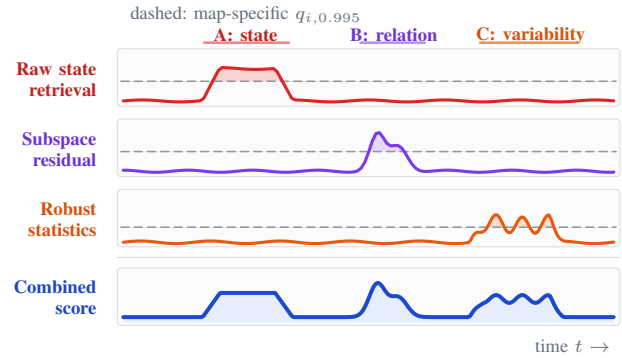
\begin{figure}[!t]
\centering
\begin{tikzpicture}[
 x=1cm,y=1cm,font=\scriptsize,
 line cap=round,line join=round,
 frame/.style={draw=inkgray!19,fill=gray!3,rounded corners=1.4pt,line width=.38pt},
 ref/.style={draw=inkgray!55,densely dashed,line width=.60pt},
 trace/.style={line width=1.17pt},
 lab/.style={font=\scriptsize\bfseries,align=right,anchor=east}
]
% Synthetic score profiles for three distinct event motifs.
% Plotted quantities are qualitative. They are not benchmark measurements.
\pgfmathdeclarefunction{axplateau}{1}{\pgfmathparse{max(0,min(1,(#1-1.05)/.23,(2.24-#1)/.23))}}
\pgfmathdeclarefunction{axpacket}{1}{\pgfmathparse{max(0,min(1,(#1-4.58)/.18,(5.82-#1)/.18))}}
\pgfmathdeclarefunction{axraw}{1}{\pgfmathparse{3.11+.013*sin(360*#1)+.43*axplateau(#1)}}
\pgfmathdeclarefunction{axsub}{1}{\pgfmathparse{2.18+.014*cos(420*#1)+.48*exp(-((#1-3.36)/.14)^2)+.33*exp(-((#1-3.64)/.17)^2)}}
\pgfmathdeclarefunction{axstats}{1}{\pgfmathparse{1.24+.017*sin(390*#1)+.24*axpacket(#1)*(1+.47*sin(1040*#1))}}
\pgfmathdeclarefunction{axfusion}{1}{\pgfmathparse{.25+.32*axplateau(#1)+.44*exp(-((#1-3.36)/.15)^2)+.26*exp(-((#1-3.64)/.16)^2)+.24*axpacket(#1)*(1+.22*sin(1040*#1))}}
% Horizontal bands: common time axis, different expert score scales.
\foreach \y in {3.04,2.11,1.18,.14} {
  \draw[frame] (1.40,\y) rectangle (8.05,{\y+.76});
}
% Event labels are intentionally different in time and morphology.
\node[font=\scriptsize\bfseries,text=rawred!88!black] at (3.10,3.98) {A: state};
\node[font=\scriptsize\bfseries,text=subpurple!90!black] at (5.06,3.98) {B: relation};
\node[font=\scriptsize\bfseries,text=statorange!92!black] at (6.87,3.98) {C: variability};
\draw[rawred!55,line width=1pt] (2.55,3.88)--(3.67,3.88);
\draw[subpurple!55,line width=1pt] (4.62,3.88)--(5.47,3.88);
\draw[statorange!65,line width=1pt] (6.20,3.88)--(7.47,3.88);
\node[lab,text=rawred!88!black] at (1.25,3.40) {Raw state\\retrieval};
\node[lab,text=subpurple!88!black] at (1.25,2.47) {Subspace\\residual};
\node[lab,text=statorange!92!black] at (1.25,1.54) {Robust\\statistics};
\node[lab,text=axonblue!90!black] at (1.25,.51) {Combined\\score};

% Colored calibration excess, separately clipped above each expert's floor.
\begin{scope}
 \clip (1.45,3.37) rectangle (8.0,3.79);
 \fill[rawred!21] (1.48,3.37) --
  plot[domain=0:6.49,samples=135,variable=\x]
    ({1.48+\x},{axraw(\x)}) -- (7.97,3.37)--cycle;
\end{scope}
\begin{scope}
 \clip (1.45,2.44) rectangle (8.0,2.86);
 \fill[subpurple!20] (1.48,2.44) --
  plot[domain=0:6.49,samples=135,variable=\x]
    ({1.48+\x},{axsub(\x)}) -- (7.97,2.44)--cycle;
\end{scope}
\begin{scope}
 \clip (1.45,1.44) rectangle (8.0,1.92);
 \fill[statorange!21] (1.48,1.44) --
  plot[domain=0:6.49,samples=220,variable=\x]
    ({1.48+\x},{axstats(\x)}) -- (7.97,1.44)--cycle;
\end{scope}
\foreach \y in {3.37,2.44,1.44} {
 \draw[ref] (1.45,\y)--(8.0,\y);
}
\draw[trace,draw=rawred]
 plot[domain=0:6.49,samples=135,variable=\x]
 ({1.48+\x},{axraw(\x)});
\draw[trace,draw=subpurple]
 plot[domain=0:6.49,samples=135,variable=\x]
 ({1.48+\x},{axsub(\x)});
\draw[trace,draw=statorange]
 plot[domain=0:6.49,samples=220,variable=\x]
 ({1.48+\x},{axstats(\x)});
% A qualitatively additive response with three separated anomaly regions.
\fill[axonblue!10] (1.48,.18) --
 plot[domain=0:6.49,samples=240,variable=\x]
 ({1.48+\x},{axfusion(\x)})--(7.97,.18)--cycle;
\draw[trace,draw=axonblue!95!black,line width=1.5pt]
 plot[domain=0:6.49,samples=240,variable=\x]
 ({1.48+\x},{axfusion(\x)});
\draw[inkgray!18,line width=.4pt] (1.40,1.04)--(8.05,1.04);
\node[anchor=west,text=inkgray!77] at (1.45,4.24)
 {dashed: map-specific $q_{i,0.995}$};
\node[anchor=north east,text=inkgray!76] at (8.05,.08)
 {time $t$ $\rightarrow$};
\end{tikzpicture}
\caption{Schematic expert-map responses to three distinct event motifs. (A) Sustained raw-retrieval excess, (B) localized subspace excess, and (C) intermittent statistics excess. Dashed lines denote map-specific normality floors, and the bottom trace illustrates their possible combination. These are not measured responses and do not establish expert specificity or diagnostic utility.}
\label{fig:teaser}
\end{figure}

\begin{figure*}[t]
\centering
\begin{tikzpicture}[
    scale=0.82,
    transform shape,
    font=\small,
    >=Stealth,
    io/.style={draw, rounded corners=3pt, thick, fill=gray!10, minimum height=0.9cm, minimum width=2.2cm, align=center, text width=1.9cm},
    proc/.style={draw, rounded corners=3pt, thick, fill=axonblueLight, draw=axonblue!80!black, minimum height=1.0cm, minimum width=2.6cm, align=center, text width=2.25cm},
    expert/.style={draw, rounded corners=3pt, thick, minimum height=1.0cm, minimum width=3.05cm, align=center, text width=2.7cm},
    cal/.style={draw, rounded corners=3pt, thick, fill=statorange!15, draw=statorange!85!black, minimum height=1.25cm, minimum width=3.15cm, align=center, text width=2.8cm},
    score/.style={draw, circle, thick, fill=shapegreen!18, draw=shapegreen!70!black, minimum size=1.15cm, align=center},
    arr/.style={->, thick, draw=inkgray}
]
    \node[io] (train) at (0,0) {Normal prefix\\$\mathbf{X}_{\mathrm{tr}}$};
    \node[proc] (norm) at (3.1,0) {Normalization\\from prefix};
    \node[proc] (period) at (6.2,0) {FFT autocorrelation\\$\to(p,W,H,R)$};
    \node[expert, fill=rawred!12, draw=rawred!80!black] (raw) at (10.0,2.0) {Raw state retrieval\\windows $8$ and $R$};
    \node[expert, fill=subpurple!12, draw=subpurple!80!black] (sub) at (10.0,0.55) {Signal/derivative\\subspace residual};
    \node[expert, fill=statorange!12, draw=statorange!80!black] (stats) at (10.0,-0.9) {Multi-scale robust\\mean and log-scale};
    \node[expert, fill=shapegreen!12, draw=shapegreen!70!black, dashed] (shape) at (10.0,-2.35) {Optional shape retrieval\\disabled in base model};
    \node[cal] (calib) at (14.4,0) {Per-map normality\\$q_{0.995}$, excess, damping};
    \node[score] (out) at (18.0,0) {$s_t$};
    \draw[arr] (train) -- (norm);
    \draw[arr] (norm) -- (period);
    \draw[arr] (period.east) -- (7.8,0);
    \draw[arr] (7.8,0) |- (raw.west);
    \draw[arr] (7.8,0) |- (sub.west);
    \draw[arr] (7.8,0) |- (stats.west);
    \draw[arr,dashed] (7.8,0) |- (shape.west);
    \draw[arr] (raw.east) -- (12.1,2.0);
    \draw[arr] (sub.east) -- (12.1,0.55);
    \draw[arr] (stats.east) -- (12.1,-0.9);
    \draw[arr,dashed] (shape.east) -- (12.1,-2.35);
    \draw[thick,draw=inkgray] (12.1,2.0) -- (12.1,-2.35);
    \draw[arr] (12.1,0) -- (calib.west);
    \draw[arr] (calib) -- (out);
\end{tikzpicture}%
\caption{The \ExCal{} pipeline. The normal prefix supplies normalization, a temporal scale from the data, exemplar memories, a cross-channel subspace, robust statistic references, and per-map normality floors. The dashed branch is implemented and evaluated as an ablation but is disabled in the current base model. The diagram describes the implemented data flow. It does not establish independent branch effects or guarantee a particular detection outcome.}
\label{fig:overview}
\end{figure*}
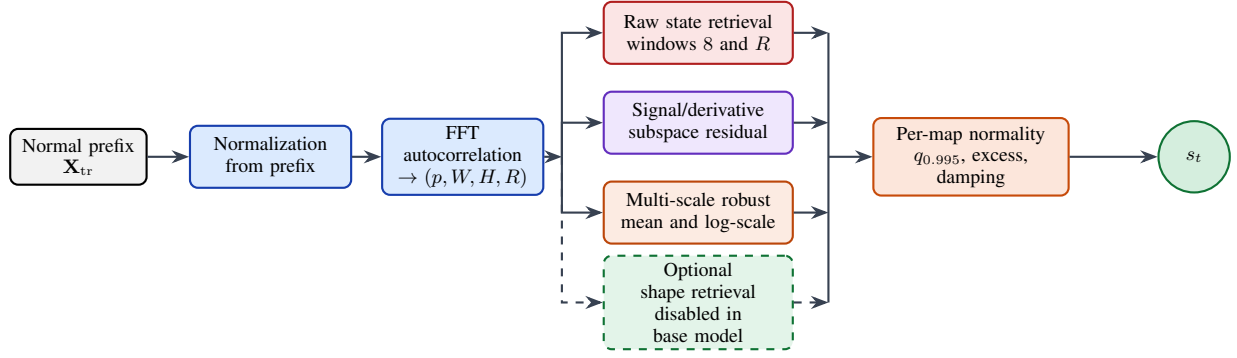

We propose \ExCal{} (\textbf{Ex}cess \textbf{Cal}ibration), which derives temporal scales and normality references from the prefix, evaluates absolute raw state, derivative-aware subspace, and multiscale statistic maps, and combines their standardized excess above expert-specific floors. Prevalence damping attenuates maps firing broadly over the scored sequence, and a small rank term preserves sub-floor ordering. An optional context-centered shape branch is evaluated but excluded from the base configuration. \cref{fig:teaser} illustrates map calibration and the resulting score, while \cref{fig:overview} details the architecture.

\ExCal{} is optimization-free: it performs no gradient updates or neural learning, but fits PCA projections, reference banks, and quantiles. The primitives are established. The contribution lies in their explicit excess-preserving calibration and fusion, evaluated using TSB-AD-M~\cite{liu_elephant_2024}, range-aware metrics~\cite{paparrizos_volume_2022}, ablations, paired tests, seed repeats, and a small timing probe. The study does not isolate causal component effects or establish generalization beyond this benchmark.

\section{Related Work}\label{sec:related}

\subsection{Learned temporal detectors}

Forecasting and reconstruction methods infer anomaly evidence from the discrepancy between an observed sequence and a learned normal model. LSTM predictors use future prediction error~\cite{malhotra_lstm_2015}. Donut models seasonal web-service KPIs with a variational autoencoder~\cite{xu_unsupervised_2018}, whereas OmniAnomaly and sequential variational models represent stochastic dynamics and temporal smoothness~\cite{su_robust_2019,li_anomaly_2021}. USAD combines autoencoding with adversarial training~\cite{audibert_usad_2020}. TranAD uses transformer encoders, self-conditioning, and adversarial learning for multivariate detection~\cite{tuli_tranad_2022}.

Attention-based methods explicitly model temporal interactions. Anomaly Transformer uses association discrepancy to distinguish anomalous points~\cite{xu_anomaly_2022}. TimesNet discovers multiple periods and maps one-dimensional variation into a two-dimensional representation~\cite{wu_timesnet_2023}. AxonAD models causal attention-query dynamics~\cite{ozer_axonad_2026}, and STREAM-VAE separates slow and fast telemetry dynamics with dual-path variational routing~\cite{ozer_stream-vae_2025}. \ExCal{} shares the objective of covering multiple temporal mechanisms, but replaces end-to-end learned fusion with explicit normality tests and calibration on the training prefix.

\subsection{Classical and representation-based detectors}

Isolation-based, density-based, histogram-based, and copula-based detectors offer simple alternatives to iterative neural training~\cite{liu_isolation_2008,hariri_extended_2021,breunig_lof_2000,goldstein_hbos_2012,li_copod_2020}. Robust principal component analysis (PCA) provides a low-rank model that tolerates sparse corruption~\cite{candes_robust_2011}. K-means and nearest-neighbor constructions supply additional geometric views of normal data~\cite{yairi_kmeans_2001,cover_knn_1967}. PyOD exposes many such detectors through a common interface~\cite{zhao_pyod_2019}. \ExCal{} follows the same transparent geometric spirit but applies its references to overlapping temporal windows and explicitly combines absolute state, derivatives, and local statistics.

\subsection{Evaluation of time-series anomalies}

The TSB-AD benchmark emphasizes that aggregate anomaly-detection results depend on dataset heterogeneity, class imbalance, and the evaluation protocol~\cite{liu_elephant_2024}. Area under the precision--recall curve (AUC-PR) and area under the receiver operating characteristic curve (AUC-ROC) evaluate score ordering without selecting one threshold, while volume under the surface for precision--recall (VUS-PR) and receiver operating characteristic (VUS-ROC) extend this idea to range-aware behavior across a volume of parameters~\cite{paparrizos_volume_2022}. We therefore treat AUC-PR and VUS-PR as the primary metrics and use the remaining reported measures as complementary diagnostics.

\section{\ExCal{}}\label{sec:method}

\subsection{Problem formulation and prefix normalization}

Let a multivariate series be
\begin{equation}
    \mathbf{X}=[\mathbf{x}_1,\ldots,\mathbf{x}_L]^{\mathsf T}\in\R^{L\times C},
\end{equation}
where $L$ is the sequence length and $C$ is the number of channels. The detector receives a normal training prefix $\mathbf{X}_{\mathrm{tr}}=\mathbf{X}_{1:L_{\mathrm{tr}}}$ and scores the full sequence. Labels are not used when fitting the detector.

Each channel is normalized with statistics computed only on the training prefix:
\begin{equation}
    z_{t,c}=
    \frac{x_{t,c}-\mu_c}{\max(\sigma_c,10^{-3})},
    \qquad
    \mu_c=\operatorname{mean}_{t\leq L_{\mathrm{tr}}}x_{t,c}.
\label{eq:normalization}
\end{equation}
This prevents the scored sequence from redefining the scale against which the normality references are calibrated. The standard-deviation floor keeps a training-constant channel numerically finite while allowing a subsequent movement to produce a large standardized deviation.

\subsection{Temporal scale from the data}

The dominant period is estimated from the mean channel autocorrelation of the first $\min(L_{\mathrm{tr}},20{,}000)$ training points. Each centered channel is transformed with a real fast Fourier transform (FFT), the inverse power spectrum produces an autocorrelation estimate, and the channel estimates are averaged. The implementation searches local maxima at lags $4$ through $\min(400,L_{\mathrm{tr}}/3)$ and accepts the candidate with the largest autocorrelation only when that autocorrelation exceeds $0.1$.

When a period $p$ is found, the implementation derives
\begin{equation}
\begin{aligned}
    W&=\clip(\lfloor2.5p\rfloor,32,512),&
    H&=\clip(\lfloor p/2\rfloor,4,64),\\
    R&=\clip(p,16,128).
\end{aligned}
\label{eq:windows}
\end{equation}
Here $W$ and $H$ define the optional context-centered shape expert, while $R$ defines the longer raw state memory in the base model. If no candidate exceeds the autocorrelation threshold, the fallback is $(W,H,R)=(100,8,32)$. The horizon is shortened when the training prefix is too short. The policy is fixed, but its effective temporal scale is derived from each normal series.

\subsection{Raw state retrieval}

The first active expert tests whether a local joint state resembles a normal state in absolute coordinates. For window length $r$,
\begin{equation}
    \mathbf{u}^{(r)}_t=
    \operatorname{vec}(\mathbf{z}_{t-r+1:t})\in\R^{rC}.
\end{equation}
The windows are projected onto a PCA basis retaining at least 95\% of sampled training window variance, and $\bar{\mathbf{u}}_r$ in \cref{eq:raw} is the mean of those sampled training-prefix windows. It is a global training reference, not a per-window local centering. The projection reduces the distance computation. Because the raw windows are not centered by their local mean, the representation retains context level. For each query window, the nearest Euclidean distance to a sampled normal exemplar bank is computed. The window score is mapped back to points by taking the maximum over windows covering each point:
\begin{equation}
    m_{\mathrm{raw}}^{(r)}(t)=
    \max_{j:\,t\in[j,j+r-1]}
    \min_{\mathbf{e}\in\mathcal{E}_r}
    \left\|
    \mathbf{V}_r^{\mathsf T}
    \left(\mathbf{u}^{(r)}_j-\bar{\mathbf{u}}_r\right)
    -\mathbf{e}
    \right\|_2.
\label{eq:raw}
\end{equation}
Two banks are used in the base model: a short window of length $8$ and the period-based window $R$. The short bank is intended to retain brief events, whereas the longer bank represents joint operating states over a broader cycle. During normality fitting, temporally overlapping self-matches are excluded whenever a non-overlapping exemplar exists.

\subsection{Optional shape retrieval with context centering}

For an additional analysis, the implementation also retains a context-centered shape expert, which is disabled in the default configuration. It centers a context and its future by the context mean, projects the context onto a 95\%-energy PCA basis, retrieves the three nearest normal contexts by cosine similarity, and averages their stored futures. The continuation error is normalized per channel and aggregated from the three largest channel values. Because centering removes local level, this branch is intended to represent continuation shape rather than absolute operating point. Its ablation is reported in Section~\ref{sec:results}.

\subsection{Signal/derivative subspace}

The implementation represents both values and first differences. \ExCal{} therefore augments every standardized point with a first difference:
\begin{equation}
    \mathbf{a}_t=
    [\mathbf{z}_t,\Delta\mathbf{z}_t]\in\R^{2C},
    \qquad
    \Delta\mathbf{z}_t=\mathbf{z}_t-\mathbf{z}_{t-1}.
\end{equation}
The covariance of the training vectors is eigendecomposed. The smallest leading basis retaining 90\% of the variance, capped at dimension 32, defines the normal subspace. The residual is
\begin{equation}
    \mathbf{r}_t=
    \left|
    (\mathbf{a}_t-\boldsymbol{\mu}_a)
    -
    \mathbf{V}_s\mathbf{V}_s^{\mathsf T}
    (\mathbf{a}_t-\boldsymbol{\mu}_a)
    \right|.
\label{eq:subspace}
\end{equation}
Each residual coordinate is robustly standardized with its training median and interquartile range. The map is the mean of the three largest standardized residual coordinates to retain the contribution of a small number of large residuals.

\subsection{Multi-scale robust statistics}

The statistics expert measures local level and variability changes at scales $s\in\{4,16,64,256\}$, subject to the available training length. Centered rolling means $\mu^{(s)}_{t,c}$ and standard deviations $\sigma^{(s)}_{t,c}$ are compared with training medians and interquartile ranges:
\begin{align}
    d^{(s)}_{\mathrm{mean},c}(t)&=
    \frac{\left|\mu^{(s)}_{t,c}-q^{(s)}_{\mathrm{mean},c,0.5}\right|}
    {\max(q^{(s)}_{\mathrm{mean},c,0.75}
    -q^{(s)}_{\mathrm{mean},c,0.25},0.05)},\\
    d^{(s)}_{\mathrm{scale},c}(t)&=
    \frac{\left|\log(\sigma^{(s)}_{t,c}+10^{-3})
    -q^{(s)}_{\log\sigma,c,0.5}\right|}
    {\max(q^{(s)}_{\log\sigma,c,0.75}
    -q^{(s)}_{\log\sigma,c,0.25},0.05)}.
\end{align}
The absolute log-scale deviation represents both variance explosions and variance collapses. The mean of the three largest channel values retains the contribution of a small number of affected channels instead of averaging across all coordinates.

\subsection{Normality floors and excess-preserving fusion}

Let $m_i(t)$ denote one map in the expert bank. During fitting, the map is evaluated on the training sequence itself. Its median and 99.5th percentile define a map-specific normality reference. At inference, only excess over the high floor is retained:
\begin{equation}
    e_i(t)=
    \left[
    \frac{m_i(t)-q_{i,0.995}}
    {\max(q_{i,0.995}-q_{i,0.5},10^{-6})}
    \right]_+.
\label{eq:excess}
\end{equation}
An observation that is ordinary for an expert contributes zero calibrated excess for that expert. This expresses different distance, residual, and statistic scales relative to their training-prefix upper spread. The floor does not guarantee a 0.5\% false-positive rate on the scored sequence.

Let $f_i$ be the fraction of points of the scored sequence with positive excess for map $i$. If a map fires broadly, its excess may reflect a broken reference or distribution shift rather than sparse anomalies. The implementation therefore applies
\begin{equation}
    h_i=\sqrt{\frac{\rho}{\max(f_i,\rho)}},
    \qquad \rho=0.10.
\label{eq:damping}
\end{equation}
Finally, a small normalized rank term preserves ordering among sub-floor points:
\begin{equation}
    s(t)=
    \sum_i\left[
    h_i e_i(t)+\tau\,\ranknorm(m_i(t))
    \right],
    \qquad \tau=0.02.
\label{eq:fusion}
\end{equation}
Here $\ranknorm(m_i(t))$ denotes the average rank of $m_i(t)$ among all points of the scored sequence, rescaled to $[0,1]$ and computed separately for each map. The rank term supplies ordering information below the floor and does not replace excess magnitude.

Both $f_i$ and $\ranknorm$ are statistics of the complete scored sequence, which under the benchmark protocol includes the training prefix. The final score is therefore a batch, transductive quantity rather than an online or causal one: the score at time $t$ depends on the distribution of map values over the entire sequence being scored, and identical points embedded in different scored sequences can receive different scores. All normal-reference quantities (normalization statistics, temporal scale, exemplar banks, subspace, statistic references, and floors) remain prefix-only. Because the prevalence and rank terms read only the unlabeled input sequence, and because every semisupervised method in the benchmark scores the same complete sequence, this is a statement about the deployment form of the detector rather than label leakage in the comparison. What this study does not evaluate is a prefix-only or causal variant of the prevalence factor and rank normalization. Such a variant is a separate design point.

\subsection{Fitting and scoring procedure}

The complete fitting and scoring path is summarized in \cref{tab:procedure}. All normal-reference quantities are constructed from the training prefix before scoring. The same expert-map code is called for the training and scored sequences. The training-map construction additionally excludes overlapping self-matches. The prevalence factors and rank normalization are then computed over the complete scored sequence, so the returned scores are batch quantities rather than prefix-only or online scores.

\begin{table}[t]
\centering
\caption{Fitting and scoring procedure for the current \ExCal{} model.}
\label{tab:procedure}
\begin{minipage}{0.96\columnwidth}
\small
\begin{enumerate}
    \item Normalize $\mathbf{X}_{\mathrm{tr}}$ and the scored sequence using the training prefix statistics.
    \item Estimate the dominant period and derive the short and period-scale windows.
    \item Build raw exemplar memories, the $[\mathbf{x},\Delta\mathbf{x}]$ subspace, and multi-scale statistic references.
    \item If enabled, build the centered-context memory and replay futures for the optional shape expert.
    \item Evaluate every active map on $\mathbf{X}_{\mathrm{tr}}$, suppress overlapping memory self-matches, and store its median and $q_{0.995}$ floor.
    \item Evaluate the same maps on the scored sequence, convert map values to calibrated excess, and compute the per-map prevalence factor over the scored sequence.
    \item Sum damped excess and sub-floor rank terms, with ranks taken over the scored sequence, to return one score per time point.
\end{enumerate}
\end{minipage}
\end{table}

\subsection{Expert score decomposition}

The final score is additive over maps, which makes an individual alert inspectable without introducing a second explanation model. For a point $t$, the contribution of map $i$ is
\begin{equation}
    s_i(t)=h_i e_i(t)+\tau\,\ranknorm(m_i(t)),
    \qquad
    s(t)=\sum_i s_i(t).
\end{equation}
The map tag identifies whether the evidence came from the short raw bank, the period-scale raw bank, the derivative-aware subspace, a rolling mean or log-scale statistic, or the optional shape branch. The floor and spread used to produce $e_i(t)$ are also available from the fitted normal reference. The decomposition identifies the normality references associated with a score. It does not establish physical root causes, diagnostic accuracy, or improvements in human operator decisions. Channel-level and domain-level diagnosis remains a separate task.

\begin{table}[t]
\centering
\caption{Fixed implementation constants in the supplied \ExCal{} configuration.}
\label{tab:constants}
\uniformtable
\begin{tabularx}{\columnwidth}{@{}lX@{}}
\toprule
\textbf{Quantity} & \textbf{Value}\\
\midrule
Exemplar and memory cap & $10{,}000$ windows\\
PCA sample cap & $4{,}096$ windows\\
Distance chunk & $1{,}024$ queries\\
Nearest neighbors & $k=3$\\
Top coordinates/channels & $3$\\
Normality floor & $q_{0.995}$\\
Prevalence knee & $\rho=0.10$\\
Sub-floor rank weight & $\tau=0.02$\\
PCA energy & $95\%$\\
Subspace energy & $90\%$\\
Statistic scales & $\{4,16,64,256\}$\\
\bottomrule
\end{tabularx}
\end{table}

\subsection{Computational profile}

The CPU implementation fits FFT-based periods, rolling statistics, PCA and subspaces, and chunked nearest-neighbor banks without backpropagation or an optimizer. Caps in \cref{tab:constants} limit reference and query batches, not total latency. Scoring is batch-oriented because prevalence and rank normalization use the full scored sequence. Online use would require prefix-only alternatives, bounded reference updates, and contamination handling. The timing probe in Section~\ref{sec:setup} covers only the current batch implementation.

\section{Experimental Methodology}\label{sec:setup}

\subsection{Benchmark and model set}

We use the official \texttt{M/eval} split of TSB-AD-M. The target contains 180 multivariate series from 17 dataset families~\cite{liu_elephant_2024}. The benchmark run contains 24 methods: eight unsupervised detectors (IForest, LOF, HBOS, KNN, KMeansAD, COPOD, EIF, and RobustPCA) and sixteen methods in the semisupervised pool (Left\_STAMPi, CNN, LSTMAD, TranAD, USAD, OmniAnomaly, PatchTST, AnomalyTransformer, TimesNet, FITS, Donut, M2N2, StreamVAE, PaAno\_PAI, AxonAD, and ExCal). The method names are retained exactly as serialized by the evaluation code.

For semisupervised methods, the training prefix is parsed from the dataset filename, the model is fitted on that prefix, and the complete sequence is scored. ExCal follows this path: only $\mathbf{X}_{\mathrm{tr}}$ determines its normalization, temporal scale, memories, subspace, statistic references, and normality floors. Its prevalence factors and rank terms are computed over the complete scored sequence, which includes the prefix, so its scores are transductive batch quantities. The benchmark labels are used only by the metric functions after scores have been generated.

\subsection{Evaluation protocol and metrics}

The supplied evaluation code loads each CSV after removing missing rows, uses all feature columns except the final label column, derives the metric sliding window with the benchmark's window-length routine at rank one on the first channel, and computes the nine reported measures:
\begin{itemize}
    \item area under the precision--recall and receiver operating characteristic curves (AUC-PR and AUC-ROC).
    \item volume under the surface for precision--recall and receiver operating characteristic (VUS-PR and VUS-ROC).
    \item standard F1 score (Standard-F1), point-adjusted F1 (PA-F1), event-based F1 (Event-F1), range-based F1 (R-F1), and affiliation F-score (Aff-F).
\end{itemize}
The VUS measures follow the benchmark metric implementation and complement conventional AUC measures for range anomalies~\cite{paparrizos_volume_2022}. The F1-style values are threshold-dependent outputs of the supplied metric routine. They should not be interpreted as performance at a separately selected deployment threshold. We report arithmetic means and standard deviations over valid series-level results rather than pooling point-level predictions.

\subsection{Seeds, hardware, and aggregation}

The benchmark uses seeds 2024, 2025, and 2026 on an Apple M3 Max MacBook Pro with 36 GB of memory. Results pool $180\times3=540$ series--seed observations for ExCal, with the same aggregation for baselines. Standard deviations reflect series heterogeneity, not confidence intervals.

All 24 methods, including every reported baseline, were rerun with the current TSB-AD evaluation code under the official split, metrics, and three-seed protocol, because the benchmark code has changed since the original repository and paper release. The ExCal row in that common rerun predates the change to the current default, in which the shape expert is disabled, so ExCal was rerun once more with the current implementation and only its row was replaced. No baseline score from the common rerun was changed. The paired statistical inputs apply the same replacement only to the ExCal column of the per-dataset matrix averaged over three seeds. This keeps the paired comparisons aligned with the implementation evaluated in this paper.

\subsection{Implementation and configuration protocol}

The constants in \cref{tab:constants} were shared across series, seeds, and variants except when a mechanism was disabled. All series-specific parameters derive from the normal prefix without labels. Aggregates, tests, and ablations use retained outputs from the supplied TSB-AD evaluation and metric code.

\subsection{Runtime benchmark}
\label{sec:runtime_protocol}

Runtime measurements were produced by the supplied benchmark timing runner. The runner evaluated all 24 benchmark methods on the first three automatically sorted TSB-AD-M files: one Genesis series with 16,220 rows and 18 channels, and two MSL series with 1,827 and 3,535 rows and 55 channels. The timing seed was 2024, with no warm-up and one measured repetition per method--series pair. The timed region starts immediately before the model-wrapper call and ends after device synchronization. CSV loading, garbage collection, cache clearing, and metric computation are excluded.

For each completed pair, the runner reports elapsed seconds and derives rows per second as the number of scored rows divided by the elapsed time. Section~\ref{sec:runtime_results} reports the median over successful cases for each method. The median is used because the three cases have different lengths and model startup costs. It is a descriptive comparison on this fixed sample, not an estimate of asymptotic complexity. The probe produced 68 successful and 4 failed records. FITS failed on all three cases because the MPS backend rejected complex-valued linear input. LSTMAD failed on the shorter MSL case. Failed records are retained as failures rather than assigned artificial timings.

\subsection{Statistical tests}

For each metric, we average the three seed results per dataset and perform paired comparisons over datasets. The omnibus test is Friedman's test across the 24 methods, with Kendall's $W$ as the effect-size measure. We then compare ExCal with each of the 23 baselines using the two-sided Wilcoxon signed-rank test after removing zero differences. The 23 resulting $p$-values per metric are adjusted with Holm's step-down procedure at $\alpha=0.05$. Affiliation-F has 175 complete datasets in the omnibus comparison because of missing baseline values. The remaining primary metric tests use all 180 datasets.

\section{Results}\label{sec:results}

\subsection{Main benchmark results}

\cref{tab:main_results} reports results for 24 methods. ExCal in its base configuration obtains AUC-PR $0.421\pm0.37$ and VUS-PR $0.452\pm0.37$, ranking second on both primary metrics behind AxonAD. It also ranks fourth on AUC-ROC and VUS-ROC, second on Standard-F1 and R-based-F1, and fifth on Affiliation-F. Point-adjusted and event-based F1 produce a different ordering. M2N2 has the highest PA-F1 and CNN has the highest Event-F1 among the listed methods.

Relative to AxonAD, ExCal is lower by $0.031$ AUC-PR and $0.041$ VUS-PR in the means pooled across seeds. It is higher than StreamVAE by $0.038$ and $0.027$, higher than PaAno\_PAI by $0.022$ and $0.047$, and higher than USAD by $0.058$ and $0.049$ on AUC-PR and VUS-PR, respectively. The standard deviations across series indicate that these averages combine different dataset behaviors rather than uniform per-series differences.

\begin{table*}[t]
\centering
\caption{TSB-AD-M results on the official \texttt{M/eval} split. Entries are mean $\pm$ standard deviation over series results pooled across seeds. The best value in each column is in bold. The second-best value is underlined. All entries have 540 valid observations for the area metrics. Affiliation-F has 537 valid observations for LOF and 528 for Left\_STAMPi. Method citations are shown beside the method names and use the supplied bibliography.}
\label{tab:main_results}
\widetable
\begin{tabularx}{\textwidth}{@{}l*{9}{Y}@{}}
\toprule
\textbf{Model} & \textbf{AUC-PR} & \textbf{AUC-ROC} & \textbf{VUS-PR} & \textbf{VUS-ROC} & \textbf{Std-F1} & \textbf{PA-F1} & \textbf{Event-F1} & \textbf{R-F1} & \textbf{Aff-F}\\
\midrule
AxonAD\textsuperscript{\cite{ozer_axonad_2026}} & $\mathbf{0.451\pm0.32}$ & $\mathbf{0.828\pm0.17}$ & $\mathbf{0.492\pm0.32}$ & $\mathbf{0.854\pm0.16}$ & $\mathbf{0.534\pm0.29}$ & $0.709\pm0.31$ & $\underline{0.615}\pm0.33$ & $\mathbf{0.483\pm0.29}$ & $0.862\pm0.13$\\
\rowcolor{axonblueLight}
\textbf{\ExCal{}} & $\underline{0.421}\pm0.37$ & $0.772\pm0.24$ & $\underline{0.452}\pm0.37$ & $0.794\pm0.24$ & $\underline{0.466}\pm0.34$ & $0.635\pm0.38$ & $0.558\pm0.38$ & $\underline{0.441}\pm0.30$ & $0.856\pm0.12$\\
PaAno\_PAI\textsuperscript{\cite{park2026paanopatchbasedrepresentationlearning}} & $0.398\pm0.34$ & $\underline{0.784}\pm0.21$ & $0.404\pm0.35$ & $0.806\pm0.21$ & $0.433\pm0.31$ & $0.773\pm0.30$ & $0.612\pm0.37$ & $0.376\pm0.27$ & $\mathbf{0.870\pm0.11}$\\
StreamVAE\textsuperscript{\cite{ozer_stream-vae_2025}} & $0.383\pm0.35$ & $0.755\pm0.24$ & $0.424\pm0.36$ & $0.771\pm0.24$ & $0.451\pm0.34$ & $0.533\pm0.38$ & $0.469\pm0.37$ & $0.417\pm0.32$ & $0.830\pm0.13$\\
USAD\textsuperscript{\cite{audibert_usad_2020}} & $0.363\pm0.34$ & $0.737\pm0.26$ & $0.403\pm0.35$ & $0.765\pm0.25$ & $0.433\pm0.32$ & $0.623\pm0.35$ & $0.519\pm0.36$ & $0.423\pm0.29$ & $0.837\pm0.13$\\
CNN\textsuperscript{\cite{ren_time-series_2019}} & $0.351\pm0.35$ & $0.774\pm0.17$ & $0.353\pm0.36$ & $\underline{0.807}\pm0.16$ & $0.413\pm0.33$ & $\underline{0.826}\pm0.27$ & $\mathbf{0.645\pm0.36}$ & $0.302\pm0.24$ & $\underline{0.867}\pm0.12$\\
TimesNet\textsuperscript{\cite{wu_timesnet_2023}} & $0.340\pm0.31$ & $0.671\pm0.30$ & $0.392\pm0.34$ & $0.711\pm0.30$ & $0.413\pm0.31$ & $0.672\pm0.32$ & $0.537\pm0.37$ & $0.408\pm0.27$ & $0.849\pm0.12$\\
PatchTST\textsuperscript{\cite{nie2023timeseriesworth64}} & $0.335\pm0.31$ & $0.675\pm0.30$ & $0.390\pm0.33$ & $0.717\pm0.30$ & $0.406\pm0.31$ & $0.708\pm0.31$ & $0.537\pm0.38$ & $0.378\pm0.25$ & $0.846\pm0.13$\\
M2N2\textsuperscript{\cite{kim2024modelmeetsnewnormals}} & $0.321\pm0.36$ & $0.741\pm0.20$ & $0.322\pm0.36$ & $0.776\pm0.19$ & $0.381\pm0.34$ & $\mathbf{0.874\pm0.19}$ & $0.601\pm0.37$ & $0.282\pm0.23$ & $0.858\pm0.12$\\
TranAD\textsuperscript{\cite{tuli_tranad_2022}} & $0.319\pm0.36$ & $0.710\pm0.23$ & $0.317\pm0.36$ & $0.745\pm0.21$ & $0.384\pm0.34$ & $0.789\pm0.32$ & $0.608\pm0.37$ & $0.284\pm0.24$ & $0.847\pm0.12$\\
OmniAnomaly\textsuperscript{\cite{su_robust_2019}} & $0.315\pm0.36$ & $0.706\pm0.22$ & $0.316\pm0.36$ & $0.744\pm0.21$ & $0.375\pm0.35$ & $0.787\pm0.32$ & $0.604\pm0.37$ & $0.294\pm0.23$ & $0.849\pm0.12$\\
FITS\textsuperscript{\cite{xu_fits_2024}} & $0.302\pm0.31$ & $0.659\pm0.30$ & $0.360\pm0.34$ & $0.709\pm0.29$ & $0.369\pm0.31$ & $0.721\pm0.31$ & $0.525\pm0.38$ & $0.280\pm0.20$ & $0.849\pm0.12$\\
Left\_STAMPi\textsuperscript{\cite{yeh_matrix_2016}} & $0.253\pm0.28$ & $0.700\pm0.21$ & $0.321\pm0.32$ & $0.735\pm0.21$ & $0.311\pm0.28$ & $0.597\pm0.36$ & $0.439\pm0.36$ & $0.302\pm0.24$ & $0.809\pm0.17$\\
KMeansAD\textsuperscript{\cite{yairi_kmeans_2001}} & $0.249\pm0.26$ & $0.687\pm0.20$ & $0.287\pm0.30$ & $0.724\pm0.20$ & $0.305\pm0.25$ & $0.672\pm0.31$ & $0.484\pm0.36$ & $0.327\pm0.23$ & $0.822\pm0.12$\\
LSTMAD\textsuperscript{\cite{malhotra_lstm_2015}} & $0.247\pm0.33$ & $0.598\pm0.34$ & $0.243\pm0.33$ & $0.623\pm0.34$ & $0.297\pm0.32$ & $0.663\pm0.41$ & $0.507\pm0.41$ & $0.201\pm0.18$ & $0.702\pm0.35$\\
RobustPCA\textsuperscript{\cite{candes_robust_2011}} & $0.239\pm0.35$ & $0.590\pm0.24$ & $0.237\pm0.35$ & $0.612\pm0.23$ & $0.283\pm0.33$ & $0.571\pm0.39$ & $0.379\pm0.38$ & $0.330\pm0.32$ & $0.790\pm0.14$\\
Donut\textsuperscript{\cite{xu_unsupervised_2018}} & $0.215\pm0.27$ & $0.621\pm0.24$ & $0.254\pm0.30$ & $0.680\pm0.24$ & $0.281\pm0.28$ & $0.527\pm0.41$ & $0.400\pm0.38$ & $0.178\pm0.17$ & $0.767\pm0.20$\\
IForest\textsuperscript{\cite{liu_isolation_2008}} & $0.211\pm0.23$ & $0.704\pm0.19$ & $0.249\pm0.26$ & $0.744\pm0.18$ & $0.281\pm0.24$ & $0.656\pm0.33$ & $0.403\pm0.32$ & $0.244\pm0.18$ & $0.801\pm0.11$\\
COPOD\textsuperscript{\cite{li_copod_2020}} & $0.205\pm0.29$ & $0.652\pm0.18$ & $0.202\pm0.29$ & $0.682\pm0.18$ & $0.269\pm0.29$ & $0.717\pm0.32$ & $0.414\pm0.36$ & $0.242\pm0.24$ & $0.799\pm0.13$\\
EIF\textsuperscript{\cite{hariri_extended_2021}} & $0.186\pm0.22$ & $0.669\pm0.17$ & $0.206\pm0.25$ & $0.706\pm0.17$ & $0.258\pm0.24$ & $0.739\pm0.29$ & $0.431\pm0.37$ & $0.257\pm0.23$ & $0.810\pm0.12$\\
HBOS\textsuperscript{\cite{goldstein_hbos_2012}} & $0.161\pm0.20$ & $0.633\pm0.18$ & $0.186\pm0.24$ & $0.669\pm0.18$ & $0.244\pm0.23$ & $0.667\pm0.33$ & $0.399\pm0.34$ & $0.244\pm0.23$ & $0.796\pm0.12$\\
KNN\textsuperscript{\cite{cover_knn_1967}} & $0.133\pm0.16$ & $0.499\pm0.22$ & $0.170\pm0.21$ & $0.572\pm0.22$ & $0.184\pm0.17$ & $0.681\pm0.37$ & $0.447\pm0.39$ & $0.205\pm0.15$ & $0.791\pm0.14$\\
LOF\textsuperscript{\cite{breunig_lof_2000}} & $0.096\pm0.09$ & $0.534\pm0.10$ & $0.132\pm0.18$ & $0.590\pm0.14$ & $0.151\pm0.12$ & $0.563\pm0.37$ & $0.325\pm0.33$ & $0.149\pm0.14$ & $0.764\pm0.13$\\
AnomalyTransformer\textsuperscript{\cite{xu_anomaly_2022}} & $0.074\pm0.07$ & $0.505\pm0.08$ & $0.117\pm0.18$ & $0.537\pm0.11$ & $0.118\pm0.10$ & $0.651\pm0.37$ & $0.396\pm0.37$ & $0.152\pm0.12$ & $0.754\pm0.18$\\
\bottomrule
\end{tabularx}
\end{table*}

\subsection{Runtime and throughput}
\label{sec:runtime_results}

\begin{figure}[t]
\centering
\begin{tikzpicture}
\begin{axis}[
    width=\columnwidth,
    height=0.68\columnwidth,
    xmode=log,
    xmin=0.002, xmax=1000,
    ymin=0.08, ymax=0.53,
    axis lines*=left,
    xlabel={Median model time per series (s, log scale)},
    ylabel={VUS-PR},
    ytick={0.1, 0.2, 0.3, 0.4, 0.5},
    xtick={0.01, 0.1, 1, 10, 100, 1000},
    grid=major,
    grid style={gray!20, thin},
    tick align=outside,
    tick style={black, thick},
    axis line style={thick},
    legend pos=south east,
    legend style={draw=none, fill=none, font=\footnotesize, cells={anchor=west}},
    clip=false
]

% Pareto front (exact coordinates)
\addplot[gray!80, densely dashed, thick, const plot] coordinates {
    (0.0082, 0.186) % HBOS
    (0.0118, 0.202) % COPOD
    (0.0662, 0.452) % ExCal
    (6.5936, 0.492) % AxonAD
    (1000.0, 0.492)
};
\addlegendentry{Pareto front}

% Semisupervised methods (14 solid blue circles, exact coordinates)
\addplot[only marks, mark=*, mark size=2.4pt, color=blue!60!cyan!80!black] coordinates {
    (6.5936,   0.492) % AxonAD
    (45.1837,  0.424) % StreamVAE
    (350.3866, 0.404) % PaAno PAI
    (1.4207,   0.403) % USAD
    (5.6295,   0.392) % TimesNet
    (9.7820,   0.390) % PatchTST
    (0.8560,   0.353) % CNN
    (0.9752,   0.322) % M2N2
    (1.7680,   0.321) % Left_STAMPi
    (2.4593,   0.317) % TranAD
    (43.6280,  0.316) % OmniAnomaly
    (4.5793,   0.254) % Donut
    (14.6543,  0.243) % LSTMAD
    (6.9608,   0.117) % AnomalyTransformer
};
\addlegendentry{Semisupervised}

% Unsupervised methods (8 hollow blue squares, exact coordinates)
\addplot[only marks, mark=square, mark size=2.3pt, thick, color=blue!60!cyan!80!black] coordinates {
    (0.0082, 0.186) % HBOS
    (0.0118, 0.202) % COPOD
    (0.0182, 0.170) % KNN
    (0.0177, 0.132) % LOF
    (0.1847, 0.287) % KMeansAD
    (0.8986, 0.206) % EIF
    (0.9919, 0.237) % RobustPCA
    (4.2834, 0.249) % IForest
};
\addlegendentry{Unsupervised}

% ExCal (red star, exact coordinates)
\addplot[
    only marks,
    mark=star,
    mark options={fill=red!75!black, draw=black, line width=0.6pt},
    mark size=5pt
] coordinates {(0.0662, 0.452)};

% Speedup arrow and label (6.5936 / 0.0662 = 99.6x)
\draw[<->, >=Stealth, thick, color=red!75!black] (axis cs:0.075, 0.475) -- (axis cs:5.9, 0.475)
    node[midway, above, font=\footnotesize\color{red!75!black}] {$\sim 100\times$ faster};

% Point labels
\node[right, font=\footnotesize\bfseries, text=red!75!black] at (axis cs:0.08, 0.452) {ExCal};
\node[above right, font=\footnotesize] at (axis cs:7.0, 0.492) {AxonAD};
\node[above, font=\footnotesize] at (axis cs:45.1837, 0.429) {StreamVAE};
\node[below, font=\footnotesize] at (axis cs:350.3866, 0.397) {PaAno PAI};
\node[left, font=\footnotesize] at (axis cs:1.25, 0.405) {USAD};
\node[left, font=\footnotesize] at (axis cs:0.75, 0.353) {CNN};
\node[right, font=\footnotesize] at (axis cs:0.21, 0.287) {KMeansAD};
\node[above, font=\footnotesize] at (axis cs:0.0118, 0.210) {COPOD};
\node[left, font=\footnotesize] at (axis cs:0.007, 0.186) {HBOS};
\node[left, font=\footnotesize] at (axis cs:5.8, 0.117) {AnomalyTransformer};

\end{axis}
\end{tikzpicture}
\caption{Accuracy-latency trade-off across 23 evaluated detectors on TSB-AD-M. The horizontal axis shows the median model-wrapper runtime per time series on a logarithmic scale, and the vertical axis shows VUS-PR. Solid circles denote semisupervised methods, and hollow squares denote unsupervised methods. The dashed gray line indicates the Pareto front. \textbf{ExCal} ($0.0662\,\text{s}$, $0.452$ VUS-PR) lies on the Pareto front and executes $99.6\times$ ($\sim 100\times$) faster than \textbf{AxonAD} ($6.5936\,\text{s}$, $0.492$ VUS-PR).}
\label{fig:runtime}
\end{figure}
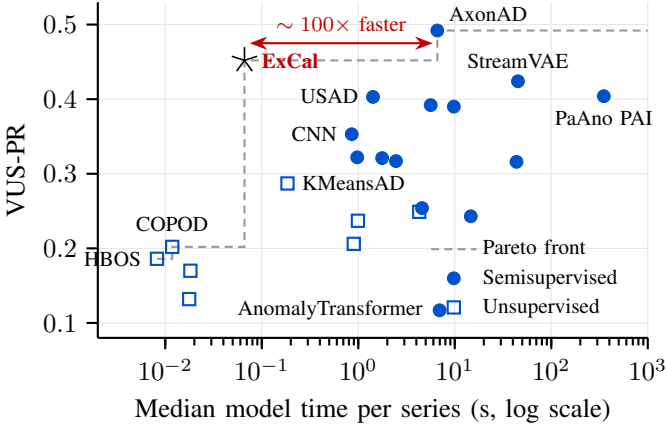

\cref{fig:runtime} reports the per-method results. The three measured ExCal times were $0.2010$ s for Genesis, $0.0486$ s for the shorter MSL series, and $0.0662$ s for the second MSL series. The median was therefore $0.0662$ s (66.2 ms), the mean was $0.1052$ s, and the median throughput was $53{,}428$ rows/s, or $2.07\times10^{6}$ values/s. ExCal was the fifth-fastest method among the 22 methods with complete timings. Its measured median was lower than that of every semisupervised comparator with three successful cases in this probe. CNN had the lowest median time among these semisupervised comparators at $0.8560$ s, about $12.9\times$ the ExCal median in this probe. These values describe the timed model-wrapper region on the three supplied cases and do not include data loading or metric computation.

\subsection{Score distributions and ranks by series}

Aggregate means do not indicate whether a method performs similarly across series. We therefore average the three executions within each dataset before computing per-series summaries. \cref{tab:per_dataset_distribution} summarizes the distribution for ExCal. The AUC-PR median is $0.3440$ with an interquartile range of $[0.0608,0.7534]$, while the VUS-PR median is $0.4108$ with an interquartile range of $[0.0665,0.8450]$. The corresponding ranges extend from $0.0029$ to $0.9999$ for AUC-PR and from $0.0032$ to $0.9998$ for VUS-PR. This spread reflects the composition of the benchmark and the per-dataset aggregation.

When methods are ranked separately on every dataset, ExCal has mean rank $7.75$ on AUC-PR and $8.09$ on VUS-PR among the 24 methods. It has rank one on 30 of 180 datasets for AUC-PR and 29 of 180 for VUS-PR, and it is in the top three on 71 and 74 datasets, respectively. The per-dataset ranks and the aggregate ranking therefore summarize different aspects of the results. ExCal has a broad but uneven rank distribution across the benchmark.

\begin{table}[t]
\centering
\caption{Per-dataset distribution of current ExCal scores after averaging the three seeds. Ranks are computed against all 24 methods for each dataset.}
\label{tab:per_dataset_distribution}
\scriptsize
\renewcommand{\arraystretch}{1.08}
\setlength{\tabcolsep}{1.2pt}
\begin{tabularx}{\columnwidth}{@{}l*{5}{Y}@{}}
\toprule
\textbf{Metric} & \textbf{Mean $\pm$ SD} & \textbf{Median} & \textbf{Q1--Q3} & \textbf{Min--Max} & \textbf{Rank / \#1 / Top-3}\\
\midrule
AUC-PR & $0.421\pm0.37$ & $0.344$ & $[0.061, 0.753]$ & $[0.003, 1.000]$ & $7.75$ / $30$ / $71$\\
VUS-PR & $0.452\pm0.37$ & $0.411$ & $[0.067, 0.845]$ & $[0.003, 1.000]$ & $8.09$ / $29$ / $74$\\
\bottomrule
\end{tabularx}
\end{table}

\subsection{Pairwise effects by series}

We report per-dataset deltas between ExCal and selected baselines in addition to aggregate means. The intervals in \cref{tab:bootstrap_effects} are nonparametric 95\% bootstrap intervals for the mean delta, based on 10,000 resamples of the 180 dataset-level seed means and a fixed bootstrap generator seed of 2026. The win--loss counts and Holm-adjusted Wilcoxon values use the same paired dataset-level observations. The bootstrap interval describes uncertainty around the arithmetic mean effect. The Wilcoxon test evaluates the signed-rank distribution. A skewed set of differences can therefore produce a positive mean without a significant majority-dataset advantage. The two viewpoints differ concretely for PaAno\_PAI on AUC-PR: the bootstrap interval contains zero, while the adjusted Wilcoxon $p$-value is $0.042$. The interval reflects uncertainty in the mean difference, whereas the test reflects the signed ranks, which favor ExCal on 106 of 180 datasets.

ExCal has a positive mean delta relative to StreamVAE, PaAno\_PAI, USAD, CNN, and AnomalyTransformer on both primary metrics. The interval for the AxonAD comparison contains zero on both metrics, and the paired tests are nonsignificant. These results do not establish equivalence. The StreamVAE intervals are positive even though ExCal wins on fewer datasets than it loses. This indicates a skewed effect distribution in which a smaller number of larger gains outweigh more numerous smaller losses. The Wilcoxon test remains nonsignificant after correction, which does not establish equivalence. Against AnomalyTransformer, ExCal has higher scores on 148 of 180 datasets for AUC-PR and 145 for VUS-PR, with positive mean deltas and small adjusted $p$-values.

\begin{table*}[t]
\centering
\caption{Per-dataset effects for selected baselines. $\Delta$ is ExCal minus the baseline after averaging seeds within each dataset. Bootstrap intervals use 10,000 resamples. W--L counts ExCal wins and baseline wins.}
\label{tab:bootstrap_effects}
\widetable
\begin{tabularx}{\textwidth}{@{}l*{8}{Y}@{}}
\toprule
\textbf{Baseline} & \multicolumn{4}{c}{\textbf{AUC-PR}} & \multicolumn{4}{c}{\textbf{VUS-PR}}\\
 & $\Delta$ & \textbf{95\% CI} & \textbf{W--L} & $p_{\mathrm{Holm}}$ & $\Delta$ & \textbf{95\% CI} & \textbf{W--L} & $p_{\mathrm{Holm}}$\\
\midrule
AxonAD\textsuperscript{\cite{ozer_axonad_2026}} & $-0.031$ & $[-0.075, 0.014]$ & 94--86 & 0.177 & $-0.041$ & $[-0.084, 0.001]$ & 92--88 & 0.276\\
StreamVAE\textsuperscript{\cite{ozer_stream-vae_2025}} & $+0.038$ & $[0.010, 0.067]$ & 88--92 & 0.111 & $+0.027$ & $[0.001, 0.055]$ & 84--96 & 0.276\\
PaAno\_PAI\textsuperscript{\cite{park2026paanopatchbasedrepresentationlearning}} & $+0.022$ & $[-0.023, 0.066]$ & 106--74 & 0.042 & $+0.047$ & $[0.009, 0.087]$ & 106--74 & 0.016\\
USAD\textsuperscript{\cite{audibert_usad_2020}} & $+0.058$ & $[0.023, 0.094]$ & 120--60 & 0.001 & $+0.049$ & $[0.014, 0.085]$ & 122--58 & 0.002\\
CNN\textsuperscript{\cite{ren_time-series_2019}} & $+0.070$ & $[0.020, 0.120]$ & 114--66 & 0.0002 & $+0.099$ & $[0.055, 0.143]$ & 113--67 & $8.94{\times}10^{-5}$\\
AnomalyTransformer\textsuperscript{\cite{xu_anomaly_2022}} & $+0.346$ & $[0.297, 0.395]$ & 148--32 & $2.54{\times}10^{-23}$ & $+0.335$ & $[0.285, 0.387]$ & 145--35 & $1.16{\times}10^{-21}$\\
\bottomrule
\end{tabularx}
\end{table*}

\subsection{Omnibus and paired tests}

The Friedman tests reject the null hypothesis of no systematic differences in method ranks for each tested metric. For AUC-PR, the statistic is $890.015$ with Kendall's $W=0.2150$. For VUS-PR it is $949.927$ with $W=0.2295$. For R-based-F1 it is $1081.997$ with $W=0.2614$. All three omnibus $p$-values are below $10^{-170}$. Affiliation-F has a smaller but still significant overall difference, with statistic $492.190$, $W=0.1223$, and $175$ complete datasets.

\cref{tab:bootstrap_effects} gives selected paired comparisons. After Holm correction, the ExCal differences from AxonAD and StreamVAE are not significant on either primary metric. The paired differences relative to PaAno\_PAI, USAD, and AnomalyTransformer are statistically significant on both primary metrics under this benchmark protocol. Across all 23 baselines, 21 comparisons reject the null for AUC-PR and the same 21 reject it for VUS-PR. The Kendall's $W$ values near $0.22$ indicate that the global ordering, although consistent across datasets, is modest in size, so these counts should be read together with the absolute effect sizes in \cref{tab:bootstrap_effects}. The tests indicate differences from most baselines. They do not establish superiority over, or equivalence to, the highest-ranked learned comparators.

\subsection{Component ablations}

\cref{tab:ablations} reports the base model and seven variants of the current implementation evaluated on the same 180 series and three seeds. Values highlighted in the table follow the reported rounded summaries. Rank-only fusion has the largest decrease relative to the base model among the evaluated variants, with changes of $-0.1042$ in AUC-PR, $-0.1050$ in VUS-PR, and $-0.1148$ in R-based-F1. Removing raw retrieval is associated with decreases of $0.0450$ in AUC-PR and $0.0471$ in VUS-PR. Removing excess calibration is associated with decreases of $0.0315$ and $0.0346$ in the two primary metrics, respectively.

Removing multi-scale statistics is associated with decreases of $0.0173$ in AUC-PR and $0.0135$ in VUS-PR. Removing the derivative subspace changes AUC-PR by only $+0.0003$ and R-based-F1 by $+0.0037$, which is consistent with overlap among the evaluated views on this benchmark. These ablations provide little evidence of an independent contribution from the derivative subspace in this configuration. Interactions among experts remain uncontrolled. Removing prevalence damping changes the primary means by less than $0.001$ and raises R-based-F1 by $0.0109$. This pattern is consistent with a larger effect on the threshold-dependent score distribution than on area-based ranking metrics, although the ablation does not isolate that mechanism.

Enabling the optional shape expert changes AUC-PR from $0.4206$ to $0.4145$ and VUS-PR from $0.4516$ to $0.4464$, while increasing AUC-ROC from $0.7722$ to $0.7858$ and VUS-ROC from $0.7935$ to $0.8038$. It also reduces R-based-F1 from $0.4407$ to $0.3769$. The default configuration therefore keeps shape retrieval disabled. This is an implementation and ablation result, not an assertion that shape information is universally unhelpful.

\begin{table*}[t]
\centering
\caption{ExCal component ablations on TSB-AD-M. Entries are mean $\pm$ standard deviation over 540 observations across series and seeds. The best value in each column is in bold. The second-best value is underlined.}
\label{tab:ablations}
\widetable
\begin{tabularx}{\textwidth}{@{}l*{9}{Y}@{}}
\toprule
\textbf{Variant} & \textbf{AUC-PR} & \textbf{AUC-ROC} & \textbf{VUS-PR} & \textbf{VUS-ROC} & \textbf{Std-F1} & \textbf{PA-F1} & \textbf{Event-F1} & \textbf{R-F1} & \textbf{Aff-F}\\
\midrule
\rowcolor{axonblueLight}
\textbf{Base, shape disabled} & $\underline{0.421}\pm0.37$ & $0.772\pm0.24$ & $\mathbf{0.452\pm0.37}$ & $0.794\pm0.24$ & $0.466\pm0.34$ & $0.635\pm0.38$ & $0.558\pm0.38$ & $0.441\pm0.30$ & $0.856\pm0.12$\\
With shape expert & $0.415\pm0.35$ & $\mathbf{0.786\pm0.24}$ & $0.446\pm0.36$ & $\mathbf{0.804\pm0.24}$ & $\mathbf{0.469\pm0.32}$ & $0.637\pm0.38$ & $0.559\pm0.38$ & $0.377\pm0.28$ & $0.842\pm0.13$\\
Without raw retrieval & $0.376\pm0.36$ & $0.733\pm0.25$ & $0.405\pm0.37$ & $0.763\pm0.24$ & $0.424\pm0.34$ & $\underline{0.687}\pm0.33$ & $0.522\pm0.38$ & $0.397\pm0.30$ & $0.841\pm0.12$\\
Without derivative subspace & $\mathbf{0.421\pm0.37}$ & $0.771\pm0.24$ & $0.450\pm0.37$ & $0.790\pm0.24$ & $\underline{0.468}\pm0.34$ & $0.624\pm0.38$ & $0.541\pm0.38$ & $\underline{0.444}\pm0.31$ & $0.853\pm0.12$\\
Without multi-scale statistics & $0.403\pm0.35$ & $\underline{0.780}\pm0.22$ & $0.438\pm0.37$ & $\underline{0.801}\pm0.22$ & $0.457\pm0.32$ & $0.667\pm0.38$ & $\mathbf{0.602\pm0.38}$ & $0.413\pm0.28$ & $\mathbf{0.862\pm0.12}$\\
Without excess calibration & $0.389\pm0.34$ & $0.759\pm0.25$ & $0.417\pm0.36$ & $0.780\pm0.25$ & $0.432\pm0.32$ & $\mathbf{0.752\pm0.30}$ & $\underline{0.588}\pm0.37$ & $0.403\pm0.29$ & $\underline{0.861}\pm0.11$\\
Without prevalence damping & $0.420\pm0.37$ & $0.772\pm0.24$ & $\underline{0.451}\pm0.37$ & $0.792\pm0.24$ & $0.467\pm0.34$ & $0.628\pm0.39$ & $0.555\pm0.38$ & $\mathbf{0.452\pm0.30}$ & $0.856\pm0.12$\\
Rank-only fusion & $0.316\pm0.33$ & $0.743\pm0.23$ & $0.347\pm0.34$ & $0.771\pm0.23$ & $0.360\pm0.31$ & $0.671\pm0.34$ & $0.507\pm0.37$ & $0.326\pm0.24$ & $0.844\pm0.11$\\
\bottomrule
\end{tabularx}
\end{table*}

\subsection{Ablation results by series}

The aggregate ablation differences are also visible at the dataset level. \cref{tab:ablation_effects} reports the mean difference between the base model and each variant after averaging seeds within each dataset. The base model has higher AUC-PR than the variant without raw retrieval on 138 of 180 datasets and higher VUS-PR on 136. For rank-only fusion, the corresponding counts are 131 and 125. By contrast, removing the derivative subspace does not produce a one-sided effect. The variant without the subspace is higher on AUC-PR for 102 datasets, while the base model is higher on VUS-PR for 83. Removing prevalence damping yields many ties because the damping factor is inactive for maps below its prevalence knee. Its aggregate effect is therefore concentrated in a smaller set of series.

The shape comparison is mixed on the area metrics. The base model has higher AUC-PR on 102 datasets and higher VUS-PR on 96, while the median per-dataset advantage is close to zero. Its R-based-F1 difference is more often positive, with 114 base wins versus 51 wins for the shape variant. These counts are consistent with using the configuration without the shape expert as the current default. They do not establish that the optional branch cannot be useful on a different data distribution.

\begin{table*}[t]
\centering
\caption{Per-dataset ablation effects after averaging the three seeds. $\Delta$ is base ExCal minus the ablation. W--L counts base wins and ablation wins, with ties omitted.}
\label{tab:ablation_effects}
\widetable
\begin{tabularx}{\textwidth}{@{}l*{6}{Y}@{}}
\toprule
\textbf{Variant} & \multicolumn{2}{c}{\textbf{AUC-PR}} & \multicolumn{2}{c}{\textbf{VUS-PR}} & \multicolumn{2}{c}{\textbf{R-based-F1}}\\
 & $\Delta$ & \textbf{W--L} & $\Delta$ & \textbf{W--L} & $\Delta$ & \textbf{W--L}\\
\midrule
With shape expert & $+0.006$ & 102--76 & $+0.005$ & 96--81 & $+0.064$ & 114--51\\
Without raw retrieval & $+0.045$ & 138--37 & $+0.047$ & 136--39 & $+0.044$ & 117--29\\
Without derivative subspace & $-0.000$ & 73--102 & $+0.001$ & 83--89 & $-0.004$ & 62--90\\
Without multi-scale statistics & $+0.017$ & 76--104 & $+0.014$ & 70--109 & $+0.028$ & 100--78\\
Without excess calibration & $+0.032$ & 118--61 & $+0.035$ & 114--63 & $+0.038$ & 119--58\\
Without prevalence damping & $+0.001$ & 55--71 & $+0.001$ & 56--66 & $-0.011$ & 42--68\\
Rank-only fusion & $+0.104$ & 131--49 & $+0.105$ & 125--55 & $+0.115$ & 126--53\\
\bottomrule
\end{tabularx}
\end{table*}

\subsection{Repeatability across seeds}

\cref{tab:seed_stability} shows the current base result separately for each execution seed. Across seeds, the mean AUC-PR varies from $0.420489$ to $0.420818$ and VUS-PR from $0.451462$ to $0.451735$. The standard deviation of the three means by seed is $1.69\times10^{-4}$ for AUC-PR and $1.39\times10^{-4}$ for VUS-PR. The corresponding 95th-percentile deviations across seeds at the individual-series level are $6.02\times10^{-4}$ and $9.67\times10^{-4}$. These are repeatability summaries, not uncertainty intervals for the benchmark population.

\begin{table}[t]
\centering
\caption{Current ExCal mean scores by execution seed.}
\label{tab:seed_stability}
\uniformtable
\begin{tabular}{lcccc}
\toprule
\textbf{Seed} & \textbf{AUC-PR} & \textbf{VUS-PR} & \textbf{AUC-ROC} & \textbf{VUS-ROC}\\
\midrule
2024 & 0.421 & 0.452 & 0.772 & 0.794\\
2025 & 0.421 & 0.452 & 0.772 & 0.794\\
2026 & 0.420 & 0.451 & 0.772 & 0.793\\
\midrule
\textbf{Pooled} & \textbf{0.421} & \textbf{0.452} & \textbf{0.772} & \textbf{0.794}\\
\bottomrule
\end{tabular}
\end{table}

\subsection{Results by dataset family}

The global standard deviation accompanies variation across the 17 source families. \cref{tab:families} shows current ExCal results by family. AUC-PR is highest on Exathlon ($0.944\pm0.12$), SWaT ($0.549\pm0.28$), SMAP ($0.491\pm0.36$), SMD ($0.483\pm0.27$), and SVDB ($0.479\pm0.29$). It is lower on GHL ($0.013\pm0.01$), Genesis ($0.004$), CreditCard ($0.054$), and OPPORTUNITY ($0.121\pm0.25$). VUS-PR is higher on Exathlon ($0.915\pm0.19$), TAO ($0.703\pm0.24$), SMAP ($0.519\pm0.37$), and SVDB ($0.515\pm0.30$), but lower on Genesis, GHL, and GECCO. These results are consistent with different performance across families. They do not establish uniform performance across all operating conditions.

\begin{table*}[t]
\centering
\caption{ExCal family-level macro results. The count is the number of unique series. Each series is first averaged over the three execution seeds, and the displayed standard deviation is then computed across series. A dash denotes a family with one series.}
\label{tab:families}
\uniformtable
\begin{tabular}{lrrrr}
\toprule
\textbf{Family} & \textbf{Series} & \textbf{AUC-PR} & \textbf{VUS-PR} & \textbf{AUC-ROC}\\
\midrule
CATSv2 & 5 & $0.157\pm0.13$ & $0.107\pm0.09$ & $0.665\pm0.17$\\
CreditCard & 1 & $0.054$ & $0.115$ & $0.764$\\
Daphnet & 1 & $0.407$ & $0.425$ & $0.944$\\
Exathlon & 25 & $0.944\pm0.12$ & $0.915\pm0.19$ & $0.965\pm0.08$\\
GECCO & 1 & $0.274$ & $0.046$ & $0.750$\\
GHL & 23 & $0.013\pm0.01$ & $0.014\pm0.01$ & $0.450\pm0.24$\\
Genesis & 1 & $0.004$ & $0.012$ & $0.595$\\
LTDB & 4 & $0.399\pm0.37$ & $0.443\pm0.38$ & $0.770\pm0.23$\\
MITDB & 11 & $0.372\pm0.27$ & $0.288\pm0.23$ & $0.811\pm0.13$\\
MSL & 14 & $0.412\pm0.39$ & $0.444\pm0.40$ & $0.831\pm0.15$\\
OPPORTUNITY & 7 & $0.121\pm0.25$ & $0.123\pm0.25$ & $0.300\pm0.31$\\
PSM & 1 & $0.212$ & $0.222$ & $0.738$\\
SMAP & 25 & $0.491\pm0.36$ & $0.519\pm0.37$ & $0.863\pm0.18$\\
SMD & 20 & $0.483\pm0.27$ & $0.442\pm0.25$ & $0.886\pm0.17$\\
SVDB & 28 & $0.479\pm0.29$ & $0.515\pm0.30$ & $0.864\pm0.13$\\
SWaT & 2 & $0.549\pm0.28$ & $0.376\pm0.04$ & $0.844\pm0.01$\\
TAO & 11 & $0.123\pm0.06$ & $0.703\pm0.24$ & $0.593\pm0.03$\\
\bottomrule
\end{tabular}
\end{table*}

\section{Discussion}\label{sec:discussion}

Ablations associate raw retrieval and excess magnitude with the largest primary-metric differences. Multiscale statistics have a smaller association, while removing the derivative subspace barely changes the primary metrics. The optional shape branch lowers AUC-PR and VUS-PR in the evaluated configuration. These are comparisons among configurations, not evidence of independent causal contributions. They also do not show that a branch is generally unhelpful.

The second-place aggregate ranking coexists with heterogeneous family results and modest Kendall's $W$ values. Corrected tests do not distinguish \ExCal{} from AxonAD or StreamVAE on either primary metric. The score decomposition identifies contributing normality maps, not physical root causes. Batch prevalence and rank statistics and a three-series, single-run timing probe preclude claims of online or real-time readiness.

\section{Limitations and Future Directions}\label{sec:limitations}

The evaluation covers one TSB-AD-M split, three seeds, and correlated performance metrics. The timing probe includes three series with one measured repetition and excludes loading, metrics, memory, and energy. A clean normal prefix is assumed. Contamination can alter reference banks and floors, while legitimate regime changes can trigger broad excess. A single detected period may miss nonstationary or multiperiodic structure. Moreover, batch prevalence and rank normalization change a point's score when the surrounding sequence changes, so a causal deployment variant remains unevaluated. Score decomposition has not been tested for operator diagnosis, and benchmark F1 values do not establish deployment thresholds. Relevant next steps include independent telemetry datasets, contamination studies, causal fusion references, threshold and detection-delay evaluation, and larger resource studies.

\section{Conclusion}\label{sec:conclusion}

\ExCal{} combines prefix-fitted raw retrieval, signal/derivative subspace residuals, and multiscale robust statistics through magnitude-preserving excess fusion without gradient optimization. Prevalence and rank terms make its inference batch and transductive. Across 180 TSB-AD-M series, the detector ranks second among 24 methods on AUC-PR and VUS-PR, although paired tests find no significant difference from AxonAD or StreamVAE. Ablations associate the largest score changes with raw retrieval and magnitude-preserving fusion. The runtime probe is preliminary, and operational performance and causal scoring remain untested.

\balance
\bibliographystyle{IEEEtran}
\bibliography{references}

% Author biographies omitted to match the IEEEtran conference format of JuRe:
% \begin{IEEEbiographynophoto}{Kadir-Kaan \"Ozer}
% is with Mercedes-Benz AG, Germany. His research interests include multivariate time-series anomaly detection, training-free machine learning, and reliable monitoring of automotive and industrial systems.
% \end{IEEEbiographynophoto}
%
% \begin{IEEEbiographynophoto}{Ren\'e Ebeling}
% is with Mercedes-Benz AG, Germany. His research interests include time-series analytics, anomaly detection, and data-driven monitoring of complex systems.
% \end{IEEEbiographynophoto}
%
% \begin{IEEEbiographynophoto}{Markus Enzweiler}
% is with the Institute for Intelligent Systems, Esslingen University of Applied Sciences, Germany. His research interests include machine learning, anomaly detection, and intelligent systems.
% \end{IEEEbiographynophoto}

\end{document}